\documentclass{article}

\usepackage{arxiv}

\usepackage[utf8]{inputenc} 
\usepackage[T1]{fontenc}    
\usepackage{hyperref}       
\usepackage{url}            
\usepackage{booktabs}       
\usepackage{amsfonts}       
\usepackage{nicefrac}       
\usepackage{microtype}      
\usepackage{lipsum}
\usepackage{graphicx}
\graphicspath{ {./images/} }
\usepackage{amsmath,amsfonts}
\usepackage{algorithmic}
\usepackage{algorithm}
\usepackage{array}
\usepackage[caption=false,font=normalsize,labelfont=sf,textfont=sf]{subfig}
\usepackage{textcomp}
\usepackage{stfloats}
\usepackage{url}
\usepackage{verbatim}
\usepackage{graphicx}
\usepackage{cite}
\usepackage{comment}
\usepackage{tablefootnote}
\usepackage{placeins}

\usepackage{svg} 

\title{PredVLA: Predictive Sensorimotor Modeling for Sub-Million-Parameter Robot Manipulation}

\author{
 Hiroki Sawada \\
  Sony Computer Science Laboratory\\
  Tokyo, Japan \\
  \texttt{hiroki.sawada@csl.sony.co.jp} \\
   \And
 Shunichi Kasahara \\
  Sony Computer Science Laboratory\\
  Tokyo, Japan \\
  \texttt{kasahara@csl.sony.co.jp} \\
}

\begin{document}
\maketitle
\begin{abstract}
Large pretrained vision-language-action models achieve strong robot-manipulation performance, while compact alternatives have largely pursued efficiency by compressing the prevailing observation-to-action paradigm.
We investigate whether predictive sensorimotor modeling can make more effective use of a limited parameter budget than direct observation-to-action mapping.
We present PredVLA, a language-conditioned predictive-coding policy with only $0.68$ million trainable network parameters and no robot-data pretraining.
Its hierarchical recurrent dynamics predict visual features and proprioception, while observations influence latent state only through prediction-error-driven online inference.
On LIBERO, PredVLA achieves an $86.9\%$ mean success rate across the three short-horizon suites and $75.4\%$ across all four suites.
Under a controlled comparison using the same frozen front end, demonstrations, action decoder, and evaluation protocol, PredVLA achieves $3.7\times$ and $7.4\times$ the three-suite mean success rates of parameter-matched Transformer and LSTM behavior-cloning policies, respectively.
A mechanism-by-mechanism transition to the recurrent behavior-cloning baseline shows that replacing the predictive pathway with direct observation input produces the largest single performance drop, accounting for approximately $70\%$ of the endpoint gap.
Further ablations identify distinct contributions from training-time latent inference, test-time error regression, hierarchical timescales, and sensory prediction-error channels.
Together, these results support predictive sensorimotor modeling as a strong inductive bias for compact language-conditioned robot control.
\end{abstract}


\section{Introduction}
\label{sec:introduction}

Vision-language-action (VLA) models have become a major approach to general-purpose robot control.
By integrating visual observations and language instructions with large-scale robot demonstration data, models such as RT-1, RT-2, Octo, OpenVLA, and $\pi_0$ have collectively demonstrated broad task coverage and transfer across objects, environments, and robot embodiments \cite{brohan2022rt,brohan2023rt,team2024octo,kim2024openvla,black2024pi_0}.
Continued advances in pretraining, fine-tuning, and action generation have pushed manipulation performance still higher, with the strongest reported LIBERO~\cite{liu2023libero} averages now exceeding $95\%$ \cite{kim2025fine,zhao2023learning,chi2025diffusion,bu2025univla}.
This progress has established large-scale pretraining as a highly effective route to language-conditioned robot manipulation.

This success, however, comes with substantial computational cost.
The strongest VLA systems contain billions of parameters \cite{kim2024openvla,black2024pi_0,bu2025univla}, making training, adaptation, and deployment expensive \cite{shukor2025smolvla}.
This has motivated a growing effort toward smaller and more efficient VLA policies.
Recent compact models such as TinyVLA, SmolVLA, and Evo-1 have shown that strong manipulation performance can be retained at substantially smaller scales \cite{wen2025tinyvla,shukor2025smolvla,lin2026evo}.
Yet these models still operate in the hundreds-of-millions parameter regime, largely pursuing efficiency by reducing the size or computational cost of architectures derived from the prevailing VLA paradigm. This raises a more fundamental question: does compact robot control require a smaller version of the same observation-to-action mapping, or can a different learning principle make substantially better use of a limited parameter budget?

We investigate prediction as such an inductive bias. Rather than feeding sensory observations directly into a policy and learning their mapping to actions, we ask whether a compact controller can instead learn the sensorimotor dynamics that generate those observations and use prediction error to infer the internal state required for action. Specifically, we test whether this predictive formulation can support strong language-conditioned manipulation with fewer than one million trainable network parameters.

Predictive coding provides one candidate for such an alternative architecture.
In predictive-coding recurrent models, internal generative dynamics predict sensory trajectories while latent state is inferred online by minimizing the discrepancy between predicted and observed signals, rather than by feeding observations directly into the recurrent dynamics \cite{ahmadi2019novel}.
This provides a recurrent formulation in which action generation and sensory feedback are mediated through learned generative dynamics and prediction-error-driven latent-state inference.

To investigate this possibility, we introduce \emph{PredVLA}, a language-conditioned predictive-coding policy that pairs a frozen multimodal front end with a trainable hierarchical recurrent generative model.
PredVLA jointly predicts visual features, proprioception, and a multimodal action distribution, while observed visual and proprioceptive signals never enter the recurrent dynamics directly.
Instead, deterministic latent variables over a recent temporal window are optimized online from sensory prediction errors together with a learned-prior complexity term, and the inferred latent trajectory is decoded into the current action.
Hierarchical time constants organize task, visual, and motor dynamics across temporal scales, while lateral pathways couple visual and action prediction.
Because visual and proprioceptive observations influence the policy only through this inference process, disabling online inference provides an exact open-loop control while preserving the learned model and language conditioning.

We evaluate PredVLA on the LIBERO benchmark using the official demonstrations and compare it against parameter-matched behavior-cloning policies implemented with a causal Transformer (BC-Transformer) and an LSTM (BC-RNN), under the same frozen front end, demonstrations, action decoder, and evaluation protocol.
With $0.68$M trainable network parameters and no robot-data pretraining, PredVLA achieves an $86.9\%$ mean success rate across the three short-horizon suites and $75.4\%$ when the long-horizon suite is included.
Its three-suite mean is $3.7\times$ and $7.4\times$ that of the parameter-matched BC-Transformer and BC-RNN, respectively.

Our contributions are as follows.

\begin{itemize}

\item We introduce \textbf{PredVLA}, a language-conditioned hierarchical predictive-coding policy with only $0.68$M trainable network parameters and no robot-data pretraining, and show that it achieves strong performance on LIBERO while substantially outperforming parameter-matched Transformer and LSTM behavior-cloning policies under a controlled evaluation protocol.

\item Through a mechanism-by-mechanism transition from PredVLA to BC-RNN, we show that the predictive formulation is a central inductive bias for strong control at this small parameter budget. Replacing the predictive pathway with direct observation input produces the largest single performance drop, accounting for approximately $70\%$ of the performance gap between PredVLA and BC-RNN.

\item We further isolate how predictive coding contributes to control: test-time error regression and training-time latent inference provide distinct gains, hierarchical timescales are particularly beneficial within the predictive architecture, and visual and proprioceptive prediction errors contribute differently across tasks. Because sensory observations influence the recurrent state only through error regression, disabling online inference additionally provides an exact open-loop counterpart without retraining or altering language conditioning.

\end{itemize}

\section{Related Work}
\label{sec:related_work}

\subsection{Efficient Vision-Language-Action Policies}

The rapid growth of VLA models has motivated a parallel effort to reduce their computational and deployment cost.
TinyVLA combines a reduced vision-language backbone with a diffusion action decoder \cite{wen2025tinyvla}.
SmolVLA reduces the model to roughly $450$M parameters through architectural simplification, layer skipping, and asynchronous inference \cite{shukor2025smolvla}.
Evo-1 operates at approximately $0.77$B parameters while seeking to preserve semantic alignment \cite{lin2026evo}.
These approaches demonstrate that substantial compression of general-purpose VLA policies is possible, but they remain in the hundreds-of-millions parameter regime.
PredVLA explores a substantially different operating point: its recurrent controller contains only $0.68$M trainable parameters, while generic visual and linguistic encoders remain frozen and no neural component is pretrained on robot data.
This separates the robot-data-trained control capacity from the generic perceptual and linguistic representations supplied by the front end.

\subsection{Test-Time Inference and Adaptation}

Several recent VLA approaches introduce online adaptation or correction during deployment.
TTT-VLA learns a latent prompt during training and optimizes that prompt at deployment through a self-supervised proxy objective while leaving the policy itself unchanged \cite{zhang2026ttt}.
RoboTTT introduces test-time-training layers whose fast weights are updated online to compress interaction history into a recurrent state \cite{jiang2026robottt}.
VLA-Corrector monitors discrepancies between predicted and observed visual dynamics and triggers corrective replanning when an open-loop action chunk begins to deviate from expectation \cite{pan2026vla}.

PredVLA differs in that the optimized quantities are latent variables already defined by the generative dynamics, while the network parameters remain fixed.
Their optimization performs online inference under the learned generative model rather than adaptation of the model parameters.
Visual and proprioceptive observations affect recurrent state only through this inference process.

\subsection{Predictive and World-Model-Based Robot Policies}

Prediction has also become increasingly prominent in large-scale robot policies.
GR-2 jointly models language, future video, and action \cite{cheang2024gr}, while Video Prediction Policy uses predictive visual representations to support action generation \cite{hu2024video}.
Unified world-model approaches couple video and action generation within shared generative models \cite{zhu2025unified,wang2026unified}, and related systems are increasingly described as world action models \cite{wang2026world,zhang2026world,liu2026oa}.
These approaches demonstrate the value of anticipating how observations and actions evolve over time.

The role of prediction in PredVLA is different.
In many world-model and video-prediction approaches, predicted future observations serve as representations, auxiliary learning targets, or variables for planning, while current observations are still supplied directly to the policy during execution.
PredVLA instead places prediction inside the online inference loop: discrepancies between predicted and observed sensory trajectories are used to revise latent state, and the inferred state in turn drives action generation.
Prediction therefore serves not only to anticipate future sensory trajectories but also as the mechanism through which new sensory evidence updates the controller's internal state.

\subsection{Predictive Coding for Sensorimotor Control}

Predictive coding provides the theoretical and computational foundation for this formulation.
In predictive-coding theories, hierarchical generative models produce top-down predictions while latent causes are revised to reduce discrepancies between predicted and observed sensory signals \cite{rao1999predictive,friston2010free}.
Under the free-energy principle, this inference can be formulated as minimizing variational free energy, which balances an accuracy term reflecting sensory prediction error against a complexity term that constrains inferred latent states toward the generative prior \cite{friston2010free}.
This provides a general framework in which internal state is inferred from prediction error under learned generative dynamics rather than determined solely by direct sensory input.

Predictive-coding recurrent models have also been extended to visuomotor learning.
Previous models jointly predicted visual and proprioceptive trajectories and inferred latent intentions through prediction-error minimization \cite{hwang2017predictive,choi2018generating}.
PV-RNN formalized online error regression over latent variables, allowing external observations to influence recurrent dynamics through backpropagated prediction errors rather than through direct forward inputs \cite{ahmadi2019novel}.
A related line of work introduced the term PC-RNN for recurrent predictive-coding architectures whose error-driven dynamics are derived from variational free-energy minimization \cite{annabi2021bidirectional,annabi2022continual}.
CERNet subsequently extended this framework to a hierarchical class-embedding PC-RNN for a physical humanoid robot, combining multi-timescale recurrent dynamics with online prediction-error-driven inference for motion generation and recognition \cite{sawada2025cernet}.
More recently, scalable predictive processing has been demonstrated with more than $30{,}000$ dimensions of visuo-proprioceptive input and multiple embodied tasks while retaining hierarchical latent dynamics \cite{idei2026predictive}.

This body of work establishes that predictive-coding recurrent models can represent high-dimensional sensorimotor dynamics and perform online latent-state inference from sensory prediction errors.
However, this line of work has not generally been evaluated as language-conditioned manipulation on standardized VLA benchmarks such as LIBERO, nor compared against contemporary policy architectures at matched trainable network parameter budgets under a shared protocol.
PredVLA addresses this gap by bringing predictive-coding recurrent inference into the modern VLA evaluation regime while operating with fewer than one million trainable network parameters.

\section{Method}
\label{sec:method}

\begin{figure*}[t]
\centering
\includegraphics[width=\linewidth]{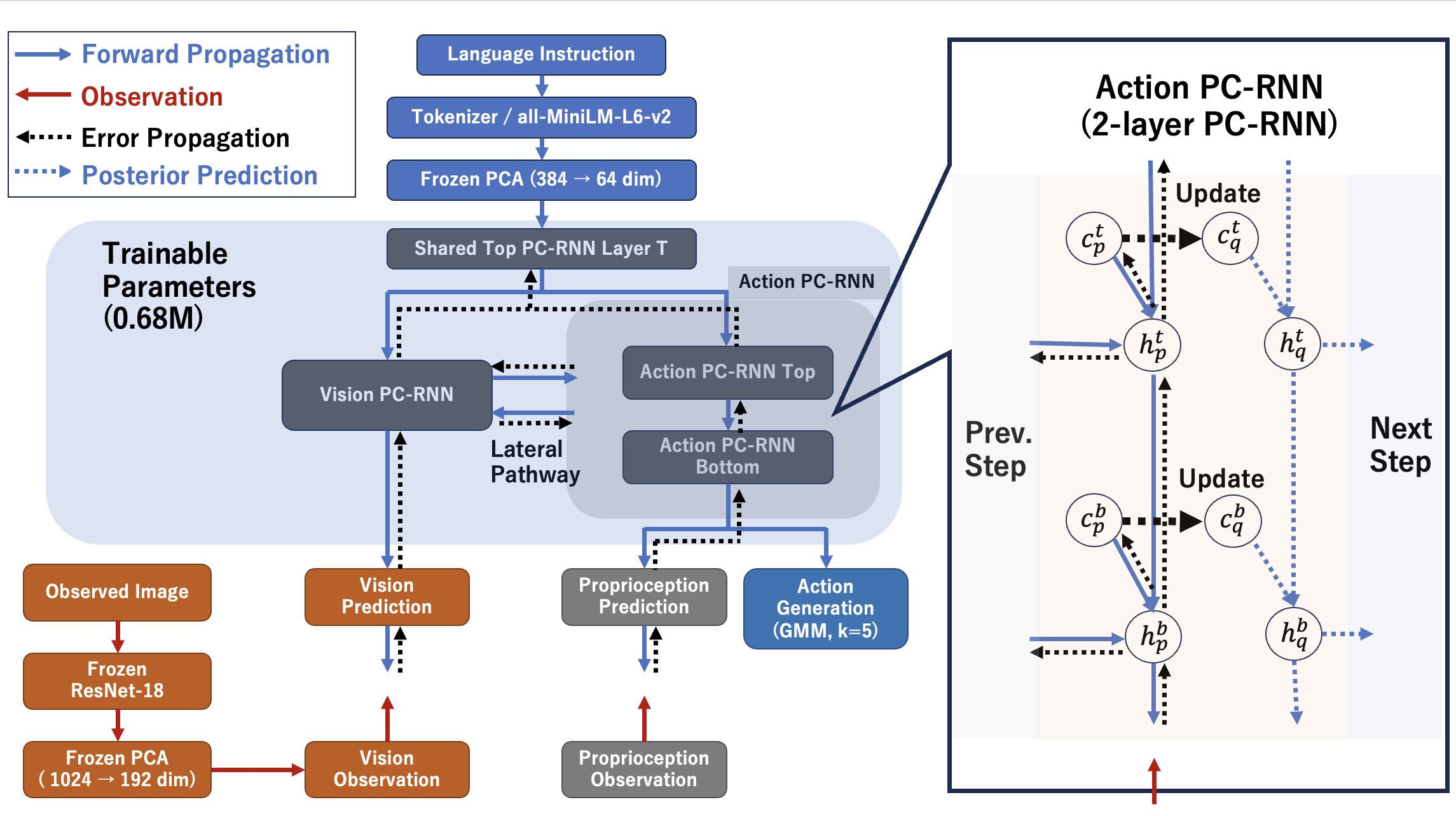}
\caption{\textbf{Structure of PredVLA.}
Blue: forward propagation.
Red: observations.
Dashed black: error propagation.
The shaded region marks the trainable recurrent controller, while the multimodal front end is frozen.
A shared top module T receives the language representation and feeds both the visual branch V and the action branch composed of A$_{\rm top}$ and A$_{\rm bottom}$.
The two branches are coupled by a visual-to-action bottleneck $V\!\to\!A$ and an efference-copy pathway $A\!\to\!V$ that carries the model's own previous action and proprioceptive predictions rather than observations.
Right: the per-step relation between the prior $c_p$, posterior $c_q$, and deterministic state $h$, shown in the conventional PV-RNN form; the prior and posterior streams are displayed side by side for contrast rather than as two concurrent recurrences.}
\label{fig:arch}
\end{figure*}

PredVLA consists of a frozen multimodal front end and a trainable hierarchical predictive-coding recurrent network (Figure~\ref{fig:arch}).
The defining property of the architecture is that sensory observations are never provided as direct inputs to the recurrent dynamics.
Instead, they appear only in the free-energy objective and influence the recurrent state through prediction-error-driven inference over latent free variables.

\subsection{Frozen front end}
\label{sec:method:frontend}

PredVLA uses fixed preprocessing for three input streams: language, vision, and proprioception.
The language instruction is embedded with the pretrained sentence encoder \texttt{all-MiniLM-L6-v2} \cite{reimers2019sentence,wang2020minilm} and projected through a fixed PCA basis, yielding the language representation $\mathbf{l}$.
The resulting representation remains constant throughout an episode.

Two camera views, a third-person \emph{agentview} and a wrist-mounted \emph{eye-in-hand} camera, are encoded independently using a frozen ImageNet-pretrained ResNet18 \cite{he2016deep}.
Each visual feature is projected through a fixed per-camera PCA basis, and the resulting representations are concatenated into the visual observation $\mathbf{v}$.
Fresh visual observations are encoded at a fixed interval during control.

Proprioception $\mathbf{q}$, consisting of joint angles and gripper state, is used directly without an encoder.

The visual and language encoders are not pretrained on robot data.
The PCA bases are fitted on the training demonstrations and then fixed throughout policy training and evaluation.

\subsection{Hierarchical dynamics}
\label{sec:method:dynamics}

The trainable network consists of recurrent modules $l\in\{\mathrm{T},\mathrm{V},\mathrm{A}_{\mathrm{top}},\mathrm{A}_{\mathrm{bottom}}\}$: a shared top module T, which serves as the root of the hierarchy, feeding a visual branch V and an action branch composed of A$_{\rm top}$ and A$_{\rm bottom}$.
T provides top-down input to both V and A$_{\rm top}$, while A$_{\rm top}$ in turn feeds A$_{\rm bottom}$.
A lateral connection from V to A$_{\rm top}$ couples the two branches within the same time step, so the modules are evaluated in the order T, V, A$_{\rm top}$, A$_{\rm bottom}$.
For the non-root modules, we denote the parent module by $\mathrm{par}(l)$, with $\mathrm{par}(\mathrm{V})=\mathrm{T}$, $\mathrm{par}(\mathrm{A}_{\mathrm{top}})=\mathrm{T}$, and $\mathrm{par}(\mathrm{A}_{\mathrm{bottom}})=\mathrm{A}_{\mathrm{top}}$.
Each module maintains a deterministic state $h^l_t\in\mathbb{R}^{d_l}$, an output $d^l_t=\tanh(h^l_t)$, and a latent free variable $c^l_t\in\mathbb{R}^{r}$.
We use a deterministic variant of the prior--posterior free-variable formulation adapted from PV-RNN \cite{ahmadi2019novel}, together with multiple-timescale recurrent dynamics \cite{yamashita2008emergence,choi2017predictive}.

For each non-root module, a top-down \emph{prior} over its free variable is generated from the output of its parent,

\begin{equation}
    c^l_{p,t}
    =
    W^l_{\rm pri} d^{\mathrm{par}(l)}_t
    +
    b^l_{\rm pri},
    \qquad
    l\neq\mathrm{T}.
    \label{eq:prior}
\end{equation}

For the root module T, the prior is generated recurrently from its own previous output,

\begin{equation}
    c^{\mathrm{T}}_{p,t}
    =
    W^{\mathrm{T}}_{\rm pri} d^{\mathrm{T}}_{t-1}
    +
    b^{\mathrm{T}}_{\rm pri}.
    \label{eq:prior_root}
\end{equation}

We denote the corresponding \emph{posterior} free variable by $c^l_{q,t}$; its online inference is described in Section~\ref{sec:method:er}.
For each non-root module, the deterministic state is updated with a module-specific time constant $\tau_l$ as

\begin{equation}
        h^l_t
        =
        \left(1-\frac{1}{\tau_l}\right)h^l_{t-1} +
        \frac{1}{\tau_l}
        \tanh\!\left(
        W^l_{\rm rec} d^l_{t-1}
        +
        W^l_{\rm td} d^{\mathrm{par}(l)}_t
        +
        U^l c^l_{q,t}
        +
        \beta^l_t
        +
        b^l
        \right),
        \qquad
        l\neq\mathrm{T}.
    \label{eq:state}
\end{equation}

The top module has no top-down state input and evolves recurrently from its own previous state while receiving the episode-level language signal,

\begin{equation}
        h^{\mathrm{T}}_t
        =
        \left(1-\frac{1}{\tau_{\mathrm{T}}}\right)h^{\mathrm{T}}_{t-1}
        +
        \frac{1}{\tau_{\mathrm{T}}}
        \tanh\!\left(
        W^{\mathrm{T}}_{\rm rec} d^{\mathrm{T}}_{t-1}
        +
        U^{\mathrm{T}} c^{\mathrm{T}}_{q,t}
        +
        W_{\rm lang}\mathbf{l}
        +
        b^{\mathrm{T}}
        \right).
    \label{eq:state_root}
\end{equation}

Here, $U^l\in\mathbb{R}^{d_l\times r}$ maps the free variable into the deterministic state space, and $\beta^l_t$ denotes a lateral input and is zero for modules without a lateral pathway.
A larger $\tau_l$ produces slower state dynamics, and the assignment of different time constants across modules defines the temporal hierarchy.

\paragraph{Language conditioning.}
The instruction is encoded once per episode into the fixed vector $\mathbf{l}$ by the frozen sentence encoder described in Section~\ref{sec:method:frontend}.
Language enters the recurrent dynamics only through the root module T via the additive term $W_{\rm lang}\mathbf{l}$ in Equation~\eqref{eq:state_root}.
Because $\mathbf{l}$ remains fixed throughout an episode, it provides an episode-level conditioning signal that propagates from T through the visual and action branches.

\paragraph{Lateral pathways.}
The visual and action branches are coupled by two lateral pathways, neither of which directly carries an observation.
Inspired by the use of low-dimensional parametric-bias representations in RNNPB \cite{tani2004self}, we introduce a low-dimensional visual-to-action bottleneck between V and A$_{\rm top}$.
Unlike a conventional parametric bias, this representation is computed dynamically from the visual state at every time step rather than represented as an independently inferred or learned bias variable.

\begin{equation}
    \begin{aligned}
        \mathbf{z}^{\mathrm{VA}}_t
        &=
        W_{\rm VA} d^{\mathrm{V}}_t,
        \\
        \beta^{\mathrm{A}_{\mathrm{top}}}_t
        &=
        W_{\rm top}\mathbf{z}^{\mathrm{VA}}_t.
    \end{aligned}
    \label{eq:visual_bottleneck}
\end{equation}

In the opposite direction, an efference-copy pathway returns the model's own previous action and proprioceptive predictions to the visual branch,

\begin{equation}
    \beta^{\mathrm{V}}_t
    =
    W_{ba}\hat{\mathbf{a}}_{t-1}
    +
    W_{bq}\hat{\mathbf{q}}_{t-1}.
    \label{eq:eff}
\end{equation}

Importantly, Equation~\eqref{eq:eff} uses predicted quantities rather than the executed action or measured proprioception, so sensory measurements do not enter the recurrent forward dynamics through this pathway.
Lateral inputs are likewise excluded from the construction of the priors in Equations~\eqref{eq:prior} and~\eqref{eq:prior_root}.

\paragraph{Decoders.}
Two-layer MLP heads decode the visual prediction $\hat{\mathbf{v}}_t$ from $d^{\mathrm{V}}_t$ and the proprioceptive prediction $\hat{\mathbf{q}}_t$ and action distribution from $d^{\mathrm{A}_{\mathrm{bottom}}}_t$.
Actions are represented by a $K$-component diagonal Gaussian mixture over the end-effector command.
The action head produces mixture weights, component means, and per-dimension scales, with the means passed through $\tanh$ and the scales constrained to remain positive.
At execution time, the mean of the highest-weight mixture component is used as $\hat{\mathbf{a}}_t$ and emitted as the current action.

\subsection{Free energy}
\label{sec:method:fe}

PredVLA performs latent-state inference by minimizing a deterministic free-energy objective derived from the accuracy--complexity formulation used in PV-RNN \cite{ahmadi2019novel}.
Visual and proprioceptive prediction errors form the accuracy terms and are averaged per dimension,

\begin{equation}
    \varepsilon_v(t)
    =
    \frac{\lVert \mathbf{v}_t-\hat{\mathbf{v}}_t\rVert^2}{\dim\mathbf{v}},
    \qquad
    \varepsilon_q(t)
    =
    \frac{\lVert \mathbf{q}_t-\hat{\mathbf{q}}_t\rVert^2}{\dim\mathbf{q}}.
    \label{eq:prediction_errors}
\end{equation}

The deviation of each posterior free variable from its corresponding generative prior forms the complexity term.
Because PredVLA uses deterministic free variables rather than stochastic latent distributions, this complexity is represented by a quadratic penalty rather than a KL divergence,

\begin{equation}
    \mathcal{C}_t
    =
    \sum_l
    w_l
    \frac{1}{2}
    \left\lVert
    c^l_{q,t}-c^l_{p,t}
    \right\rVert^2.
    \label{eq:comp}
\end{equation}

For a temporal index set $\mathcal{I}$, the free energy is therefore expressed as

\begin{equation}
    E(\mathcal{I})
    =
    \sum_{t\in\mathcal{I}}
    \left[
    \lambda_v\varepsilon_v(t)
    +
    \lambda_q\varepsilon_q(t)
    +
    \mathcal{C}_t
    \right].
    \label{eq:fe}
\end{equation}

The index set $\mathcal{I}$ denotes the sequence over which the objective is evaluated: the training sequence during learning and the recent sliding window during online error regression.
Here, $\lambda_v$ and $\lambda_q$ weight the visual and proprioceptive prediction errors, respectively.
Prediction-error terms are masked when the corresponding observation is unavailable and over padded portions of a sequence.

\subsection{Training}
\label{sec:method:train}

During training, the posterior free variables are model parameters, with one $c^l_{q,t}$ for each module, time step, and demonstration, and are optimized jointly with the network weights by backpropagation through time.
The training objective augments the free-energy terms with the likelihood of the demonstrated action,

\begin{equation}
    \mathcal{L}
    =
    \lambda_v\bar{\varepsilon}_v
    +
    \lambda_q\bar{\varepsilon}_q
    +
    \lambda_a\overline{\mathrm{NLL}}(\mathbf{a})
    +
    \mathcal{C},
    \label{eq:train}
\end{equation}

where $\overline{\cdot}$ denotes a mask-normalised mean over the sequence and $\mathrm{NLL}$ is the negative log-likelihood of the demonstrated action under the Gaussian-mixture action distribution.
The action-likelihood term is used only for training and is not included in the free energy minimized during test-time inference.
The posterior free variables $c^l_{q,t}$ are optimized during training but are not part of the deployed controller and are discarded after training.
Concrete optimization and sequence settings are reported in Section~\ref{sec:setup}.

\subsection{Inference by error regression}
\label{sec:method:er}

At test time, the network weights are frozen and only the posterior free variables are updated.
Inference minimizes the free energy in Equation~\eqref{eq:fe} over a sliding temporal window of recent observations.
At each control step we

\begin{enumerate}
\item append the current observation to the inference window and initialize the posterior free variables of the newest step from their corresponding priors in Equations~\eqref{eq:prior} and~\eqref{eq:prior_root};
\item perform $n_{\rm itr}$ optimization steps on the posterior free variables, $c_q \leftarrow c_q-\eta\,\partial E/\partial c_q$, regenerating the deterministic recurrent states over the window after each update;
\item decode the current action from $d^{\mathrm{A}_{\mathrm{bottom}}}$.
\end{enumerate}

Because the deterministic states are regenerated from the inferred free variables rather than edited directly, each state trajectory remains consistent with the learned recurrent dynamics.
The concrete inference-window length, optimizer, number of iterations, step size, and complexity weighting used in the experiments are reported in Section~\ref{sec:setup}.

\paragraph{Exact open-loop control.}
Setting $n_{\rm itr}=0$ disables error regression entirely.
The posterior free variables then remain equal to their priors, which are generated solely from the recurrent hierarchy, and sensory observations appear nowhere in the recurrent forward computation.
The emitted actions are therefore independent of visual and proprioceptive observations by construction while the learned network and language conditioning remain unchanged.
This provides an exact open-loop counterpart to the closed-loop policy without retraining or modifying the model.

\section{Experimental setup}
\label{sec:setup}

\subsection{Benchmark and data}

We evaluate PredVLA on the LIBERO benchmark \cite{liu2023libero}, using the \textsc{spatial}, \textsc{goal}, \textsc{object}, and \textsc{long} (\texttt{libero\_10}) suites. \textsc{spatial}, \textsc{goal}, and \textsc{object} each contain ten short-horizon tasks that vary, respectively, in object placement, goal predicate, and object identity. \textsc{long} contains ten longer-horizon tasks, with demonstrations averaging $276$ steps compared with $125$--$149$ steps for the three short-horizon suites. Each suite provides $500$ official demonstrations, with $50$ demonstrations per task. Each PredVLA controller is trained on a single suite and evaluated on the corresponding suite.

The PCA bases used by the frozen front end are fitted before policy training and remain fixed thereafter. In this experiment, shared PCA bases are fitted jointly on the training demonstrations from all four suites and used for every suite. The same preprocessing is applied to PredVLA and all parameter-matched baselines.

\subsection{Model configuration}
\label{sec:setup:model}

Table~\ref{tab:model_config} summarizes the concrete architectural settings used in the experiments. The recurrent architecture itself is defined in Section~\ref{sec:method:dynamics}; the values reported here specify the front-end dimensions, recurrent dimensions, output distribution, and temporal hierarchy used for LIBERO. The three short-horizon suites use the same model configuration. For \textsc{long}, the recurrent time constants are widened while the remaining architecture is unchanged.

\begin{table}[hbtp]
\centering
\caption{PredVLA model configuration used.}
\label{tab:model_config}
\small
\begin{tabular}{@{}p{0.61\linewidth}p{0.29\linewidth}@{}}
\toprule
Setting & Value \\
\midrule
Language encoder output / PCA dimension & $384 \rightarrow 64$ \\
Visual encoder output / PCA dimension per camera & $512 \rightarrow 96$ \\
Number of camera views & $2$ \\
Concatenated visual dimension & $192$ \\
Visual refresh interval & $4$ control steps \\
Proprioceptive dimension & $9$ \\
Action dimension & $7$ \\
\midrule
Top-module state dimension $d_{\mathrm{T}}$ & $64$ \\
Visual-module state dimension $d_{\mathrm{V}}$ & $256$ \\
A$_{\rm top}$ state dimension $d_{\mathrm{A}_{\mathrm{top}}}$ & $256$ \\
A$_{\rm bottom}$ state dimension $d_{\mathrm{A}_{\mathrm{bottom}}}$ & $256$ \\
Free-variable dimension $r$ & $64$ per module \\
Language projection $W_{\rm lang}$ & $64\times64$, no bias \\
Visual-to-action bottleneck & $256\rightarrow32\rightarrow256$, no bias \\
Action-to-visual projection & $7\rightarrow256$ \\
Proprioception-to-visual projection & $9\rightarrow256$ \\
\midrule
Gaussian-mixture components $K$ & $5$ \\
Action-mean activation & $\tanh$ \\
Action-scale activation & softplus \\
Minimum action scale & $0.05$ \\
\midrule
Short-horizon $\tau$ (T, V, A$_{\rm top}$, A$_{\rm bottom}$) & $(16,8,5,2)$ \\
Long-horizon $\tau$ (T, V, A$_{\rm top}$, A$_{\rm bottom}$) & $(30,14,8,2)$ \\
Trainable controller parameters & $675{,}732$ \\
\bottomrule
\end{tabular}
\end{table}

\subsection{Training and inference configuration}
\label{sec:setup:optimization}

Table~\ref{tab:optimization_config} summarizes the optimization settings used for PredVLA. All suites use the same optimizer and loss weights, while the training sequence length differs between the short- and long-horizon suites.

Visual prediction errors are evaluated only on control steps for which a fresh visual observation is available, and all loss terms are masked over padded portions of the training sequences. At test time, the module-specific complexity weights used during training are replaced by the shared inference weight $w_{\rm ER}$. The same inference configuration is used for all four suites.

\begin{table}[hbtp]
\centering
\caption{PredVLA training and online-inference configuration.}
\label{tab:optimization_config}
\small
\begin{tabular}{@{}p{0.61\linewidth}p{0.29\linewidth}@{}}
\toprule
Setting & Value \\
\midrule
\multicolumn{2}{@{}l}{\textit{Training}} \\
Training steps & $30{,}000$ \\
Batch size & $32$ \\
Optimizer & Adam \\
Network learning rate & $6\times10^{-5}$ \\
Posterior-variable learning rate & $10^{-2}$ \\
Learning-rate schedule & constant \\
Weight decay & $10^{-4}$, network weights only \\
Gradient clipping & enabled \\
Short-horizon sequence length & $200$ steps \\
Long-horizon sequence length & $500$ steps \\
Visual prediction weight $\lambda_v$ & $1$ \\
Proprioceptive prediction weight $\lambda_q$ & $1$ \\
Action-likelihood weight $\lambda_a$ & $1$ \\
Complexity weights $(w_{\mathrm{T}},w_{\mathrm{V}},w_{\mathrm{A}_{\mathrm{top}}},w_{\mathrm{A}_{\mathrm{bottom}}})$ & $(0.05,0.02,0.02,0.01)$ \\
\midrule
\multicolumn{2}{@{}l}{\textit{Online error regression}} \\
Optimizer & Adam \\
Inference window $W$ & $40$ steps \\
Inference iterations $n_{\rm itr}$ & $10$ \\
Inference step size $\eta$ & $0.05$ \\
Inference complexity weight $w_{\rm ER}$ & $1.0$ for all modules \\
\bottomrule
\end{tabular}
\end{table}

\subsection{Evaluation protocol}

Success is determined by the benchmark's task-specific success predicate. Rollouts start from the official evaluation initial states, which are separate from the demonstration initial states.

For the main evaluation, we run $50$ rollouts per task on each LIBERO suite. With ten tasks per suite and $14$ independently trained seeds, this corresponds to $500$ episodes per seed and $7{,}000$ episodes per suite. Each rollout is capped at $600$ control steps. We report both the mean across the three short-horizon suites and the mean across all four suites.

Ablation experiments use $20$ rollouts per task and $7$ seeds per condition. Architectural ablations are retrained for each condition, whereas inference-time ablations reuse the trained checkpoints and modify only the specified inference setting.

To examine how online inference depends on each sensory channel, we additionally perturb the visual and proprioceptive observations used for error regression at test time. For one channel at a time, a fraction $\alpha\in\{0.25,0.50,0.75,1.00\}$ of the observations in the inference window is replaced by observations from an unrelated episode, while the other channel is left unchanged. These evaluations use the same $20$ rollouts per task and $7$ seeds as the other inference-time ablations.

\subsection{Baselines}

All baselines use the same frozen front end, training demonstrations, Gaussian-mixture action head, and evaluation protocol as PredVLA. We compare against a causal Transformer and an LSTM whose trainable network parameter counts are matched to PredVLA within $5\%$, together with action-chunking Transformer variants using chunk lengths of $8$ and $16$.

The baselines are trained for $50{,}000$ steps, compared with $30{,}000$ steps for PredVLA, and their learning rates are selected from a four-point sweep. The action-chunking variants use ACT-style temporal ensembling \cite{zhao2023learning}. Their trainable network parameter counts are $1.14\times$ and $1.36\times$ that of PredVLA for chunk lengths $8$ and $16$, respectively.

\subsection{From PredVLA to BC-RNN: a mechanism-by-mechanism ladder}
\label{sec:ladder}

The controlled baseline comparison measures the performance gap between PredVLA and conventional behavior-cloning policies, but the two endpoints differ in several architectural and inference mechanisms simultaneously. To decompose this gap, we construct a mechanism-by-mechanism ladder from PredVLA to the BC-RNN baseline. Starting from the intact controller, each rung removes or replaces one mechanism while preserving the modifications introduced by the preceding rungs. This produces a sequence of intermediate policies between predictive coding and conventional recurrent behavior cloning. We additionally evaluate one side branch, L2b, to test whether the effect of removing latent inference can be attributed simply to the absence of a direct visual input to the action pathway.

\paragraph{Ladder construction.}

\begin{description}
\item[L0 --- PredVLA.] The intact model, including latent free variables $c$, the accuracy--complexity free-energy objective, hierarchical recurrent dynamics, and online error regression with $n_{\rm itr}=10$.

\item[L1 --- no online error regression.] The same trained PredVLA controller is evaluated with $n_{\rm itr}=0$. No network parameters are retrained or otherwise modified. Because sensory observations enter the recurrent dynamics only through error regression, this yields the exact open-loop counterpart of L0.

\item[L2 --- no free variables or latent free-energy inference.] The free variables $c$ and their quadratic complexity penalty are removed. The controller retains the hierarchical, multi-timescale predictive recurrent architecture and is trained using the sensory-prediction and action-likelihood objectives, but no latent variables are optimized during either training or evaluation. All rungs from L2 onward therefore operate without online error regression.

\item[L2b --- direct visual input to the action pathway.] As a side branch from L2, the frozen visual features are additionally provided through a direct feedforward pathway to the lower action module. This condition tests whether the performance change at L2 is explained by visual information reaching the action pathway only indirectly through the learned predictive dynamics. L2b is not used as a step in the main L0--L6 chain.

\item[L3 --- no sensory prediction.] Starting from L2, the sensory-prediction pathway is removed and current observations are instead supplied directly to the recurrent policy during forward computation. The controller therefore retains recurrent state and the remaining hierarchical organization, but no longer uses prediction as the mechanism linking sensory observations to action generation.

\item[L4 --- no temporal hierarchy.] Starting from L3, all recurrent time constants are set to $\tau=1$. This removes the layer-specific temporal integration $(16,8,5,2)$ used by PredVLA and reduces the recurrent dynamics to single-timescale updates.

\item[L5 --- LSTM recurrence.] The recurrent update used in L4 is replaced by LSTM recurrence while the remaining ladder inputs and connections are retained. This isolates the effect of replacing the predictive-coding-derived recurrent dynamics with a conventional gated recurrent cell.

\item[L6 --- BC-RNN.] The endpoint of the ladder is the parameter-matched LSTM behavior-cloning baseline used in the controlled baseline comparison. Its architecture and optimization protocol are unchanged from the baseline experiment described above.
\end{description}

\paragraph{Training and parameter matching.}

The newly constructed architectural variants L2, L2b, L3, L4, and L5 are trained from scratch for $30{,}000$ steps using the same training demonstrations, batch size, Adam optimizer, constant network learning rate of $6\times10^{-5}$, and other applicable optimization settings as PredVLA. Terms or parameters made inapplicable by a particular rung are simply removed. L0 and L1 share the same trained PredVLA checkpoints, while L6 reuses the existing BC-RNN baseline trained according to the $50{,}000$-step baseline protocol above.

For the main ladder, recurrent width is adjusted where necessary so that the trainable network parameter count remains within $1\%$ of PredVLA's $675{,}732$ parameters. The resulting recurrent widths are $d=256$ for L2, $d=301$ for L3 and L4, and $d=142$ for L5. L6 contains $674{,}276$ trainable network parameters. The sole exception is the L2b side branch: adding the direct visual pathway increases its parameter count to $724{,}884$, or $7.3\%$ above PredVLA. Thus, any performance decrease in this condition cannot be attributed to a smaller parameter budget.

\paragraph{Evaluation protocol.}

The ladder is evaluated on the three short-horizon suites (\textsc{spatial}, \textsc{goal}, and \textsc{object}). Each evaluated suite--seed combination consists of ten tasks with $50$ rollouts per task, yielding $500$ episodes per seed. Seven independently trained seeds are used for each newly trained ladder condition.

The five newly trained variants L2, L2b, L3, L4, and L5 therefore comprise $5\times3\times7=105$ independently trained suite--seed models and $52{,}500$ evaluation episodes in total. L0 and L1 reuse PredVLA checkpoints, and L6 reuses the independently trained BC-RNN baseline. Where differences from the intact model are reported, they are computed using the matched seven-seed subset so that seed-level differences can be paired; the $14$-seed result remains the primary estimate reported for PredVLA in the main evaluation.

\section{Results}
\label{sec:results}

\subsection{Main results}
\label{sec:results:main}

Table~\ref{tab:main} reports the performance of PredVLA on the four LIBERO suites. With $675{,}732$ trainable network parameters and no robot-data pretraining, PredVLA achieves success rates of $83.19\pm6.10$ on \textsc{spatial}, $88.40\pm3.83$ on \textsc{goal}, $89.24\pm6.13$ on \textsc{object}, and $40.57\pm6.44$ on \textsc{long}. The mean success rate is $86.94$ across the three short-horizon suites and $75.35$ across all four suites.

\begin{table*}[htbp]
\centering
\small
\caption{PredVLA performance on LIBERO. Results are mean success rates (\%) $\pm$ standard deviation across $14$ independently trained seeds.}
\label{tab:main}
\begin{tabular}{lcccccc}
\toprule
Model & \textsc{spatial} & \textsc{goal} & \textsc{object} & 3-suite mean & \textsc{long} & 4-suite mean \\
\midrule
PredVLA & $83.19\pm6.10$ & $88.40\pm3.83$ & $89.24\pm6.13$ & $\mathbf{86.94}$ & $40.57\pm6.44$ & $\mathbf{75.35}$ \\
\bottomrule
\end{tabular}
\end{table*}

\paragraph{Comparison with controlled baselines.}

Table~\ref{tab:baselines} compares PredVLA with behavior-cloning policies trained using the same frozen front end, demonstrations, action representation, and evaluation protocol. Across all four suites, PredVLA achieves a mean success rate of $75.35$, compared with $19.73$ for the parameter-matched Transformer and $10.26$ for the parameter-matched LSTM. The action-chunking Transformer variants achieve $16.70$ and $3.44$ with chunk lengths of $8$ and $16$, respectively. PredVLA runs at an end-to-end control latency of $46$\,ms per step ($21.6$\,Hz) on an RTX~5090, including the frozen front end and online error regression.

\begin{table*}[hbtp]
\centering
\small
\caption{Controlled baseline comparison.}
\label{tab:baselines}
\begin{tabular}{llccccccc}
\toprule
& Model & Params & vs.\ ours & \textsc{spatial} & \textsc{goal} & \textsc{object} & \textsc{long} & Mean \\
\midrule
B1 & BC-Transformer & $646{,}628$ & $0.957\times$ & $24.89\pm4.19$ & $30.83\pm5.98$ & $15.49\pm12.15$ & $7.71\pm2.43$ & $19.73$ \\
B2 & BC-RNN (LSTM) & $674{,}276$ & $0.998\times$ & $15.89\pm4.25$ & $16.94\pm3.42$ & $2.20\pm2.27$ & $6.00\pm6.00$ & $10.26$ \\
B3 & BC-TF, chunk $8$ & $772{,}558$ & $1.143\times$ & $21.11\pm3.78$ & $32.57\pm2.98$ & $12.83\pm13.64$ & $0.29\pm0.76$ & $16.70$ \\
B4 & BC-TF, chunk $16$ & $916{,}478$ & $1.356\times$ & $3.77\pm1.33$ & $9.40\pm2.67$ & $0.31\pm0.66$ & $0.29\pm0.76$ & $3.44$ \\
\midrule
-- & \textbf{PredVLA (ours)} & $\mathbf{675{,}732}$ & $1.000\times$ & $\mathbf{83.19}$ & $\mathbf{88.40}$ & $\mathbf{89.24}$ & $\mathbf{40.57}$ & $\mathbf{75.35}$ \\
\bottomrule
\end{tabular}
\end{table*}

\paragraph{Comparison with published models.}

\begin{figure}[hbtp]
\centering
\includegraphics[width=0.8\linewidth]{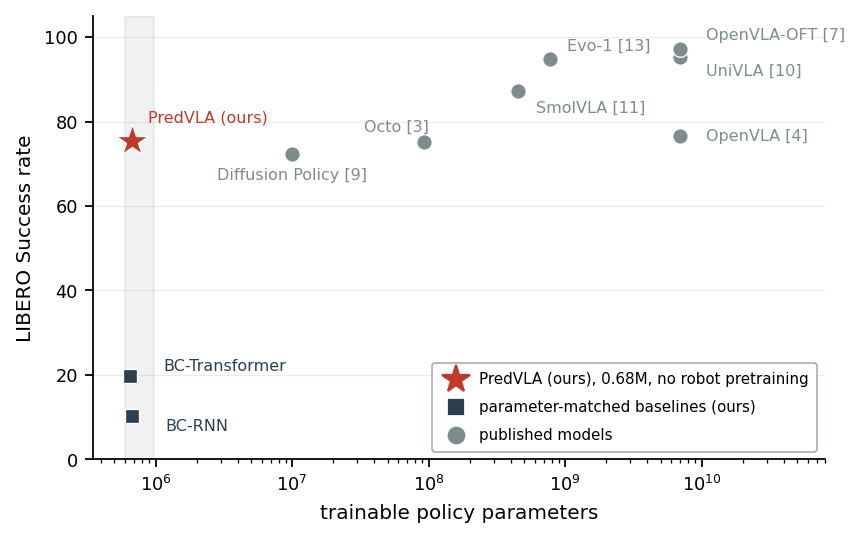}
\caption{Trainable policy parameters versus LIBERO benchmark mean success rate. Shaded region indicates the parameter range of the controlled baselines in Table~\ref{tab:baselines}.}
\label{fig:params}
\end{figure}

Figure~\ref{fig:params} places PredVLA among published LIBERO results as a function of trainable policy parameters. PredVLA reaches a four-suite mean of $75.35$ with $0.68$M trainable network parameters. For reference, OpenVLA \cite{kim2024openvla} reports $76.50$ with a $7$B-parameter model, while the OpenVLA evaluation of Octo \cite{team2024octo} reports $75.08$ with $93$M parameters. These published results differ in pretraining, optimization, and evaluation protocol and are included only to provide context for the parameter scale; Table~\ref{tab:baselines} provides the controlled comparison.

\subsection{From PredVLA to BC-RNN: a mechanism-by-mechanism ladder}
\label{sec:results:ladder}

\begin{figure*}[htbp]
\centering
\includegraphics[width=0.82\linewidth]{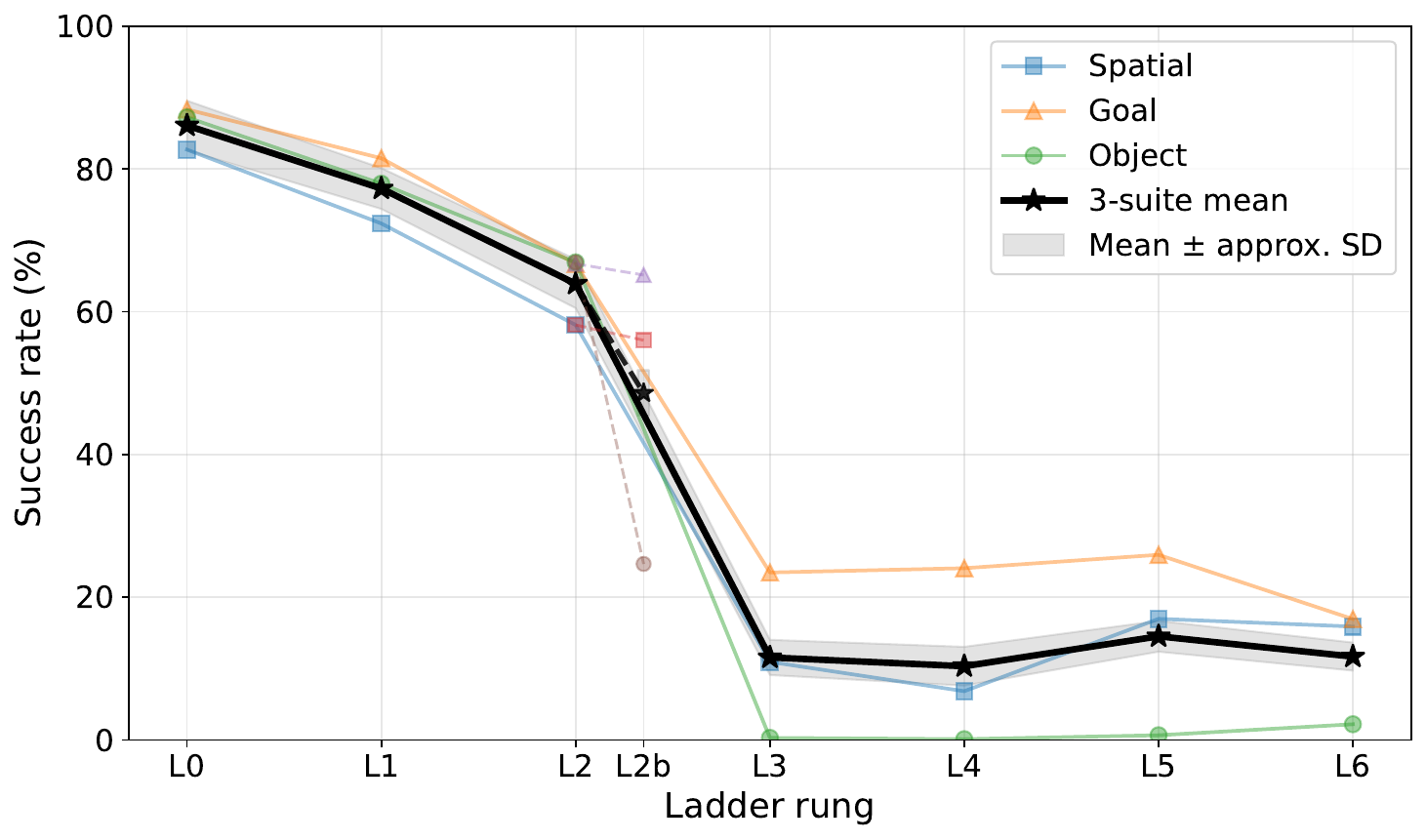}
\caption{Performance across the mechanism-by-mechanism ladder from PredVLA (L0) to BC-RNN (L6). Thin lines show success rates on \textsc{spatial}, \textsc{goal}, and \textsc{object}, while the thick black line shows their three-suite mean. Each successive rung removes or replaces one component of the preceding model. The dominant performance decrease occurs when sensory prediction is removed (L2$\rightarrow$L3), after which flattening the temporal hierarchy and replacing the recurrent cell produce comparatively small changes. L2b is a side branch from L2 in which visual features are additionally supplied directly to the action pathway; dashed segments distinguish this branch from the main ladder.}
\label{fig:ladder}
\end{figure*}

Table~\ref{tab:ladder} traces the performance gap between PredVLA and BC-RNN by sequentially removing or replacing mechanisms, as defined in Section~\ref{sec:ladder}. Figure~\ref{fig:ladder} visualizes the corresponding progression for each suite and for the three-suite mean.

\begin{table*}[hbtp]
\centering
\small
\caption{The ladder from PredVLA to BC-RNN. Success rates are mean (\%) $\pm$ standard deviation across seven seeds. $\Delta$ is the change in the three-suite mean relative to the preceding rung. L2b is a side branch from L2 and is not part of the main chain.}
\label{tab:ladder}
\begin{tabular}{llrrrrrr}
\toprule
& Modification & Params & \textsc{spatial} & \textsc{goal} & \textsc{object} & Mean & $\Delta$ \\
\midrule
L0 & --- (intact PredVLA) & 675{,}732 & $82.71 \pm 6.50$ & $88.34 \pm 4.00$ & $87.23 \pm 7.10$ & 86.09 & --- \\
L1 & $-$ Online error regression & 675{,}732 & $72.36 \pm 6.00$ & $81.50 \pm 3.67$ & $77.93 \pm 4.94$ & 77.26 & $-8.83$ \\
L2 & $-$ Predictive Coding Inference & 675{,}732 & $58.14 \pm 5.89$ & $66.71 \pm 6.42$ & $66.91 \pm 5.42$ & 63.92 & $-13.34$ \\
L3 & $-$ Prediction Pathway & 674{,}560 & $10.94 \pm 3.96$ & $23.46 \pm 6.23$ & $\phantom{0}0.29 \pm 0.60$ & 11.56 & $\mathbf{-52.36}$ \\
L4 & $-$ Temporal hierarchy ($\tau{=}1$) & 674{,}560 & $\phantom{0}6.83 \pm 3.88$ & $24.06 \pm 7.17$ & $\phantom{0}0.11 \pm 0.16$ & 10.33 & $-1.23$ \\
L5 & PV-RNN cell $\to$ LSTM & 676{,}928 & $16.97 \pm 3.96$ & $25.94 \pm 5.05$ & $\phantom{0}0.66 \pm 0.91$ & 14.52 & $+4.19$ \\
L6 & Existing BC-RNN & 674{,}276 & $15.89 \pm 4.25$ & $16.94 \pm 3.42$ & $\phantom{0}2.20 \pm 2.27$ & 11.68 & $-2.84$ \\
\midrule
L2b & L2 $+$ Direct visual input & 724{,}884 & $56.00 \pm 5.99$ & $65.14 \pm 6.16$ & $24.66 \pm 4.30$ & 48.60 & $-15.32$ \\
\bottomrule
\end{tabular}
\end{table*}

\paragraph{Reading the ladder.}

First, the ladder approaches the independently implemented BC-RNN endpoint. After removing latent inference and prediction, flattening the temporal dynamics, and replacing the recurrent cell with an LSTM, L5 reaches a three-suite mean of $14.52$, compared with $11.68$ for L6. The aggregate difference is only $2.84$ points, and the \textsc{spatial} scores differ by $1.08$ points ($16.97$ versus $15.89$). Although the suite-level correspondence is not exact, particularly on \textsc{goal}, the low-performance regime of the independently implemented BC-RNN is closely reproduced by the sequence of controlled architectural changes. The ladder therefore connects PredVLA to BC-RNN through explicit architectural changes that progressively recover the observed performance gap.

Second, the largest single change occurs when prediction is removed. The transition from L2 to L3 reduces the three-suite mean by $52.36$ points, accounting for $70.4\%$ of the $74.41$-point difference between L0 and L6. As also evident in Figure~\ref{fig:ladder}, all three suites collapse toward the BC-RNN performance regime at this transition. Once prediction has been removed, subsequently setting $\tau=1$ changes the mean by only $-1.23$ points, and replacing the recurrent cell with an LSTM changes it by $+4.19$ points. This contrasts with the substantial effect of removing the temporal hierarchy from the intact predictive architecture in Table~\ref{tab:ablation}, where success decreases by approximately $17$--$29$ points across the three short-horizon suites. Although the two temporal-hierarchy manipulations are not identical, the combined results suggest a strong interaction: hierarchical timescales contribute substantially within the predictive architecture but provide little benefit after the predictive mechanism itself has been removed.

Third, test-time inference and the learned latent-inference mechanism make distinct contributions. Disabling online error regression while keeping the trained network fixed (L0 $\to$ L1) reduces the three-suite mean by $8.83$ points. Removing the free variables and latent free-energy inference during training (L1 $\to$ L2) produces a further $13.34$-point reduction. Across these two consecutive rungs, the components associated with free-variable inference account for $22.17$ points, or approximately $30\%$ of the total L0--L6 gap. The effect of the free-variable formulation is therefore not confined to test-time optimization; the training-time formulation also contributes substantially to the resulting policy.

The L2b side branch tests whether the L2 degradation can be explained simply by the absence of a direct visual shortcut to the action pathway. Adding such a feedforward visual pathway does not recover performance: the three-suite mean decreases from $63.92$ in L2 to $48.60$ in L2b. The largest change occurs on \textsc{object}, where success falls from $66.91$ to $24.66$, despite L2b having $724{,}884$ trainable parameters and therefore being the largest model in the ladder. Thus, the loss at L2 cannot be explained simply by insufficient direct access to visual features; providing an additional feedforward visual input is not a substitute for the predictive pathway in this architecture.

\paragraph{Limitations.}

The final transition from L5 to L6 is not a controlled architectural edit: L6 is the existing BC-RNN baseline, trained independently under the baseline optimization protocol rather than as another rung within the ladder implementation. The close aggregate performance of L5 and L6 should therefore be interpreted as an endpoint consistency check rather than as evidence that the two implementations are identical. In addition, the ladder is evaluated only on the three short-horizon suites and does not establish whether the same decomposition holds for \textsc{long}.

\subsection{Ablation Study}

Table~\ref{tab:ablation} evaluates the main architectural and inference components of PredVLA. A1--A3 modify the architecture and are retrained for each condition, whereas A4--A6 modify only test-time inference. Because seed-to-seed variability on LIBERO is substantial, we report paired differences between the intact model and each ablation using matched seeds.

\paragraph{Architectural components.}

Replacing the hierarchical time constants with a common value reduces success by $18.66$, $16.99$, and $29.03$ percentage points on \textsc{spatial}, \textsc{goal}, and \textsc{object}, respectively. Among the architectural ablations, this is the only modification that consistently degrades performance across all three short-horizon suites, with the largest drop occurring on \textsc{object}.

Removing the visual-to-action bottleneck reduces \textsc{spatial} and \textsc{goal} by $24.23$ and $16.06$ points, respectively, but has essentially no effect on \textsc{object} ($+0.04$). Removing the efference-copy pathway has a weaker and more suite-dependent effect, with reductions of $6.51$, $11.84$, and $3.31$ points on \textsc{spatial}, \textsc{goal}, and \textsc{object}, respectively. Together, these results indicate that the temporal hierarchy contributes broadly across suites, whereas the two lateral pathways provide more suite-dependent benefits.

\paragraph{Online error regression.}

Disabling online error regression reduces success by $6.44$, $9.56$, and $12.17$ points on \textsc{spatial}, \textsc{goal}, and \textsc{object}, respectively, and by $12.57$ points on \textsc{long}. Because visual and proprioceptive observations affect the recurrent dynamics only through error regression, setting $n_{\rm itr}=0$ provides an exact open-loop condition without changing the learned network or language conditioning. The corresponding open-loop success rates are $73.79$, $78.50$, $78.14$, and $28.00$ on \textsc{spatial}, \textsc{goal}, \textsc{object}, and \textsc{long}, respectively. Thus, online error regression improves closed-loop execution across all four suites in the evaluated conditions.

\paragraph{Sensory error channels.}

A4 and A5 separately remove the visual and proprioceptive prediction-error terms from online inference. Their effects are strongly suite dependent. On \textsc{spatial}, removing the visual error term reduces success by $3.51$ points, compared with $2.80$ points when removing the proprioceptive term. On \textsc{object}, the pattern is markedly different: removing the proprioceptive term reduces success by $14.17$ points, whereas removing the visual term reduces success by only $2.96$ points. On \textsc{goal}, removing the visual and proprioceptive terms changes success by $-4.99$ and $-1.41$ points, respectively.

The strongest asymmetry therefore appears on \textsc{object}. Notably, the mean reduction caused by removing only the proprioceptive error term ($-14.17$) is numerically larger than that caused by disabling error regression entirely ($-12.17$). These ablations were not directly compared, so this difference should be interpreted descriptively; nevertheless, it suggests that incomplete sensory correction can in some cases be less effective than fully open-loop execution.

\begin{figure*}[htbp]
\centering
\includegraphics[width=\linewidth]{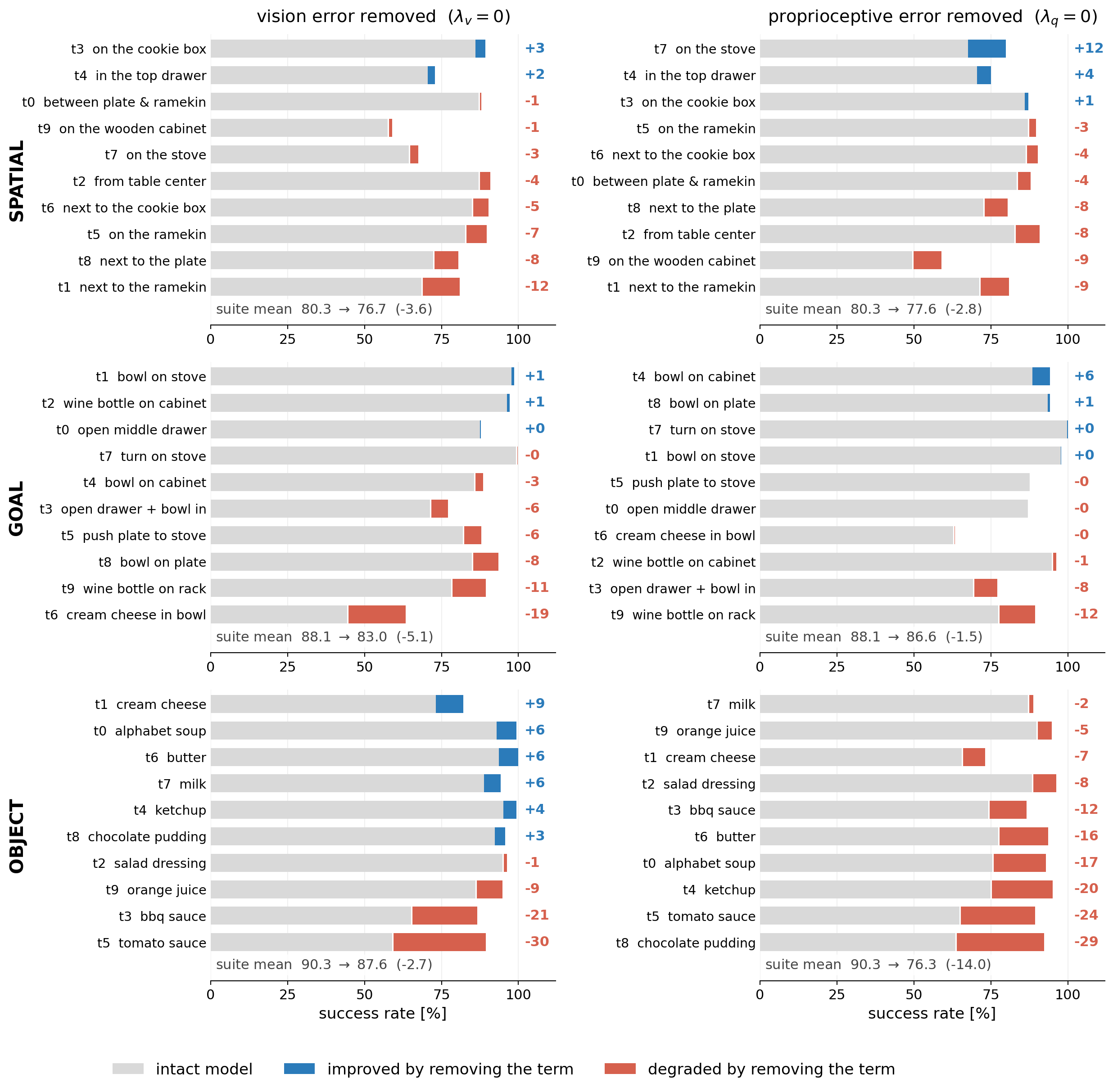}
\caption{Task-level effects of removing the visual prediction-error term ($\lambda_v=0$) or the proprioceptive prediction-error term ($\lambda_q=0$) during online inference. Within each suite, tasks are sorted by the success-rate change induced by the ablation. Gray bars indicate the intact model, while colored bars indicate the change after removing the corresponding error term. Although the suite-level averages in Table~\ref{tab:ablation} are similar for some suites, the task-level patterns differ substantially, especially on \textsc{object}, where performance is much more sensitive to the proprioceptive prediction-error term.}
\label{fig:ablation_pertask}
\end{figure*}

Figure~\ref{fig:ablation_pertask} further shows that the suite-level averages hide substantial task-level variation. On \textsc{spatial} and \textsc{goal}, the visual and proprioceptive ablations often affect overlapping sets of tasks and yield relatively similar suite-level effects. On \textsc{object}, however, the distinction is much clearer: removing the visual term leaves most tasks only mildly affected and improves six of the ten tasks, whereas removing the proprioceptive term causes large drops on several tasks. Thus, the contribution of each sensory error channel is neither uniform across suites nor across tasks, and aggregate success rates alone can obscure which source of prediction error drives online correction.

\begin{table*}[hbtp]
\centering
\small
\caption{Ablation results. A1--A3 require retraining, whereas A4--A6 are inference-time changes applied to trained checkpoints. $\Delta$ denotes the mean paired difference from the intact model evaluated using the same seven seeds.}
\label{tab:ablation}
\begin{tabular}{llcccc}
\toprule
& Change from PredVLA & \textsc{spatial} & \textsc{goal} & \textsc{object} & \textsc{long} \\
\midrule
A1 & $-$ Temporal hierarchy ($\tau$ set to its mean) & $61.57\pm8.73$ & $71.07\pm12.09$ & $61.29\pm12.99$ & --- \\
& \quad $\Delta$ & $-18.66$ & $-16.99$ & $-29.03$ & --- \\
A2 & $-$ Visual-to-action bottleneck & $56.00\pm15.64$ & $72.00\pm2.75$ & $90.36\pm3.36$ & --- \\
& \quad $\Delta$ & $-24.23$ & $-16.06$ & $+0.04$ & --- \\
A3 & $-$ Efference copy A$\to$V & $73.71\pm8.87$ & $76.21\pm6.18$ & $87.00\pm6.58$ & --- \\
& \quad $\Delta$ & $-6.51$ & $-11.84$ & $-3.31$ & --- \\
A4 & $-$ Visual error in ER ($\lambda_v{=}0$) & $76.71\pm5.72$ & $83.07\pm7.25$ & $87.36\pm5.34$ & --- \\
& \quad $\Delta$ & $-3.51$ & $-4.99$ & $-2.96$ & --- \\
A5 & $-$ Proprioceptive error in ER ($\lambda_q{=}0$) & $77.43\pm7.47$ & $86.64\pm6.87$ & $76.14\pm10.16$ & --- \\
& \quad $\Delta$ & $-2.80$ & $-1.41$ & $-14.17$ & --- \\
A6 & $-$ Online ER ($n_{\rm itr}{=}0$) & $73.79\pm4.38$ & $78.50\pm9.14$ & $78.14\pm4.53$ & $28.00\pm6.62$ \\
& \quad $\Delta$ & $-6.44$ & $-9.56$ & $-12.17$ & $-12.57$ \\
\bottomrule
\end{tabular}
\end{table*}

\FloatBarrier

\paragraph{Sensitivity to corrupted observations.}

\begin{figure*}[htbp]
\centering
\includegraphics[width=\linewidth]{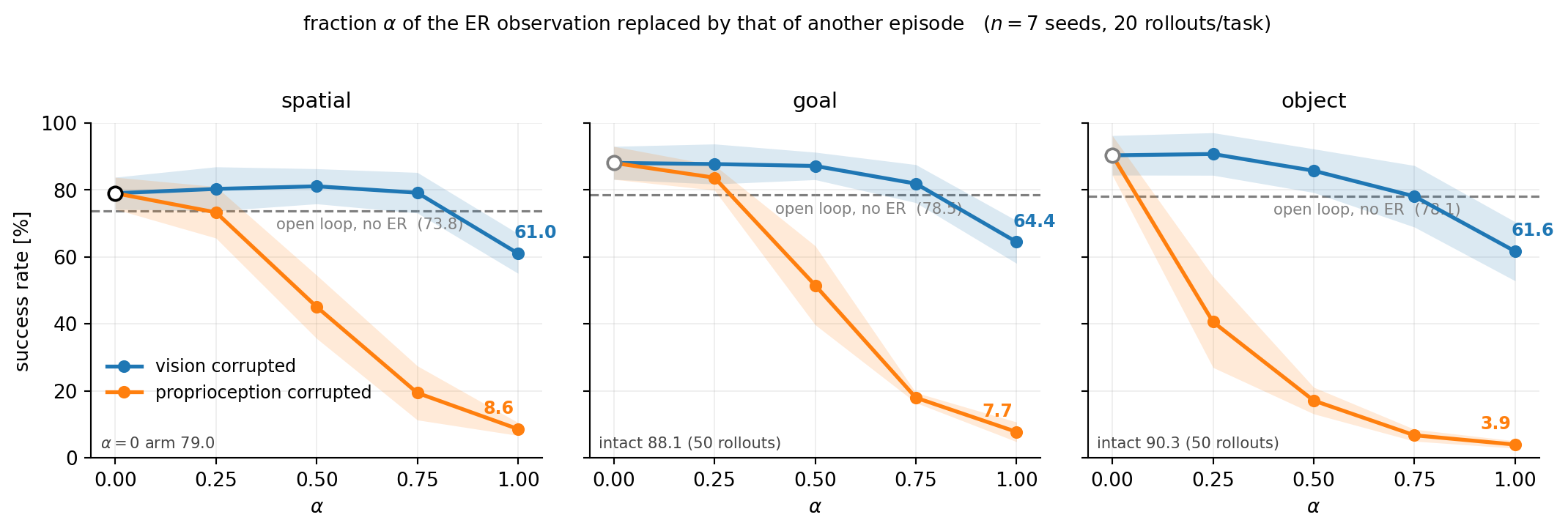}
\caption{Effect of corrupting the visual or proprioceptive observations used for error regression. $\alpha$ denotes the fraction of observations replaced by those from an unrelated episode. Bands indicate $\pm1$ seed standard deviation, and the dashed line indicates the corresponding open-loop score.}
\label{fig:redundancy}
\end{figure*}

Visual and proprioceptive corruption produce substantially different effects (Figure~\ref{fig:redundancy}). At $\alpha=0.50$, corrupting the visual observations changes success by less than five points on all three short-horizon suites. In contrast, proprioceptive corruption at only $\alpha=0.25$ reduces success by $5.71$, $4.41$, and $49.74$ points on \textsc{spatial}, \textsc{goal}, and \textsc{object}, respectively. Proprioceptive corruption remains more detrimental than visual corruption as the corruption level increases.

\section{Discussion}
\label{sec:discussion}

\subsection{Prediction as an inductive bias for compact control}

The main result of this study is not only that strong language-conditioned manipulation can be achieved with fewer than one million trainable network parameters, but that the predictive formulation appears to be central to achieving this performance at such a small parameter budget.
PredVLA reaches an $86.94\%$ mean success rate across the three short-horizon LIBERO suites and $75.35\%$ across all four suites with $675{,}732$ trainable network parameters and no robot-data pretraining.
Under the controlled comparison, parameter-matched Transformer and LSTM behavior-cloning policies achieve substantially lower performance despite sharing the same frozen front end, demonstrations, action representation, and evaluation protocol.

The mechanism-by-mechanism ladder provides further evidence that this difference is associated with how sensory information is incorporated into the controller rather than simply with the use of recurrent dynamics or a particular recurrent cell.
Replacing the predictive pathway with direct observation input reduces the three-suite mean from $63.92\%$ to $11.56\%$, a $52.36$-point decrease that accounts for $70.4\%$ of the total performance difference between PredVLA and BC-RNN.
After this transition, removing the temporal hierarchy and replacing the recurrent update with an LSTM produce comparatively small changes, and the resulting policy approaches the independently implemented BC-RNN endpoint.
Together, these results suggest that predictive sensorimotor modeling provides a strong inductive bias for extracting useful control structure from a limited parameter budget.

This interpretation does not imply that predictive coding is generally superior to Transformer- or LSTM-based policies.
Larger pretrained VLA models achieve substantially higher performance under different model scales, datasets, and optimization regimes, and increasing the capacity of the controlled baselines here would move the comparison outside the sub-million-parameter setting studied in this work.
Rather, the results indicate that strong manipulation performance at this operating point does not require a large robot-data-trained observation-to-action policy and that changing the computational formulation of control can be at least as important as increasing model capacity.

\subsection{Latent inference, online correction, and temporal hierarchy}

The ladder further separates the contribution of the predictive formulation from that of prediction-error-driven inference.
Disabling online error regression while keeping the trained controller fixed reduces the three-suite mean by $8.83$ points, whereas additionally removing the free variables and latent free-energy inference during training produces a further $13.34$-point decrease.
Thus, the contribution of latent inference is not confined to test-time optimization: the training-time formulation also affects the policy that is ultimately learned.

Because observations enter the recurrent dynamics only through error regression, setting $n_{\rm itr}=0$ provides an exact open-loop counterpart using the same learned controller and language conditioning.
This allows the contribution of online sensory correction to be isolated without retraining or introducing a separately trained open-loop model.
The relatively high open-loop performance on the three short-horizon suites indicates that much of their behavior can be generated from the learned recurrent dynamics alone, while the larger reduction on \textsc{long} suggests a greater dependence on online correction as task horizon and error accumulation increase.

The results also indicate that temporal hierarchy is most useful within the predictive architecture rather than as an isolated architectural feature.
Replacing the hierarchical time constants with a common value in the intact predictive controller reduces success by approximately $17$--$29$ percentage points across the three short-horizon suites.
In contrast, once the predictive pathway has been removed in the ladder, subsequently setting $\tau=1$ changes the three-suite mean by only $-1.23$ points.
Although these two manipulations are not identical, their combined pattern suggests that hierarchical timescales and predictive dynamics are strongly complementary: temporal abstraction provides substantial benefit when it structures a learned predictive process, but little additional benefit once that predictive process has been removed.

\subsection{Sensory error channels}

The visual and proprioceptive prediction errors do not contribute uniformly across tasks.
Removing either term individually has relatively modest effects on \textsc{spatial} and \textsc{goal}, whereas \textsc{object} is substantially more sensitive to removal of the proprioceptive error term.
The task-level analysis in Figure~\ref{fig:ablation_pertask} further shows that similar suite-level averages can conceal markedly different effects across individual tasks.

The corrupted-observation experiments reveal a related asymmetry.
Partial corruption of the visual observations has only a small effect over a broad range, whereas proprioceptive corruption produces substantially larger performance losses, particularly on \textsc{object}.
These results suggest that prediction-error-driven inference does not depend on a fixed sensory channel but can exploit different error signals depending on the structure of the task.
In particular, the strong sensitivity of \textsc{object} to proprioceptive corruption suggests that accurate estimation of the robot's own interaction state can be more important than visual correction for some manipulation behaviors.

Because these experiments perturb the observations available to the inference process rather than the physical environment itself, they should not be interpreted as evidence of general robustness to environmental disturbances.
More broadly, the present results are limited to simulation, a single robot embodiment, and separately trained controllers for each LIBERO suite.
Whether the same predictive inductive bias remains effective under physical disturbances, broader multi-task training, and substantially longer-horizon behavior remains an important question for future work.

\section{Conclusion}
\label{sec:conclusion}

We presented PredVLA, a language-conditioned predictive-coding policy with $0.68$M trainable network parameters and no robot-data pretraining.
On LIBERO, PredVLA achieves a mean success rate of $86.94\%$ across the three short-horizon suites and $75.35\%$ across all four suites, substantially outperforming parameter-matched Transformer and LSTM behavior-cloning policies under a controlled evaluation protocol.
A mechanism-by-mechanism transition from PredVLA to BC-RNN further shows that replacing the predictive pathway with direct observation input produces the largest single performance drop, accounting for approximately $70\%$ of the performance gap between the two endpoints.
Together, these results support predictive sensorimotor modeling as a strong inductive bias for achieving language-conditioned manipulation with a small trainable controller.

Further analyses show that this advantage is not attributable to a single component of the predictive-coding formulation.
Training-time latent inference and test-time error regression provide distinct contributions, while hierarchical timescales are particularly beneficial within the predictive architecture.
Because sensory observations affect the recurrent dynamics only through prediction-error-driven inference, disabling online error regression additionally provides an exact open-loop counterpart without retraining or altering language conditioning.
Visual and proprioceptive prediction errors contribute differently across tasks, further indicating that the inference process can exploit different sensory signals depending on the manipulation problem.

The present study is limited to simulation, a single robot embodiment, and the nominal LIBERO task distribution, with a separate controller trained for each suite.
The substantially lower performance on \textsc{long} also shows that longer-horizon manipulation remains a challenge for the current architecture.
Future work will examine whether the same predictive inductive bias extends to physical-robot deployment, environmental perturbations, broader multi-task training, and longer-horizon behavior while retaining a small trainable parameter budget.

\section*{Acknowledgment}
This work was supported by JST CREST Grant Number JPMJCR2552.

\bibliographystyle{IEEEtran}
\bibliography{references}

@article{brohan2022rt,
  title={Rt-1: Robotics transformer for real-world control at scale},
  author={Brohan, Anthony and Brown, Noah and Carbajal, Justice and Chebotar, Yevgen and Dabis, Joseph and Finn, Chelsea and Gopalakrishnan, Keerthana and Hausman, Karol and Herzog, Alex and Hsu, Jasmine and others},
  journal={arXiv preprint arXiv:2212.06817},
  year={2022}
}

@article{zhang2026ttt,
  title={TTT-VLA: Test-Time Latent Prompt Optimization for Vision-Language-Action Models},
  author={Zhang, Wenbo and Li, Jianxiong and Yang, Shuai and Chen, Sijin and Liu, Jiajun and Liu, Lingqiao and Ma, Xiao},
  journal={arXiv preprint arXiv:2606.03127},
  year={2026}
}

@article{jiang2026robottt,
  title={RoboTTT: Context Scaling for Robot Policies},
  author={Jiang, Yunfan and Chebotar, Yevgen and Zheng, Ruijie and Hu, Fengyuan and Ge, Yunhao and Wu, Jimmy and Dai, Tianyuan and Reed, Scott and Fei-Fei, Li and Zhu, Yuke and others},
  journal={arXiv preprint arXiv:2607.15275},
  year={2026}
}

@article{tani2004self,
  title={Self-organization of distributedly represented multiple behavior schemata in a mirror system: reviews of robot experiments using RNNPB},
  author={Tani, Jun and Ito, Masato and Sugita, Yuuya},
  journal={Neural Networks},
  volume={17},
  number={8-9},
  pages={1273--1289},
  year={2004},
  publisher={Elsevier}
}

@inproceedings{wang2020minilm,
  title={MiniLM: Deep Self-Attention Distillation for Task-Agnostic Compression of Pre-Trained Transformers},
  author={Wang, Wenhui and Wei, Furu and Dong, Li and Bao, Hangbo and Yang, Nan and Zhou, Ming},
  booktitle={Advances in Neural Information Processing Systems},
  volume={33},
  pages={5776--5788},
  year={2020}
}

@inproceedings{reimers2019sentence,
  title={Sentence-bert: Sentence embeddings using siamese bert-networks},
  author={Reimers, Nils and Gurevych, Iryna},
  booktitle={Proceedings of the 2019 conference on empirical methods in natural language processing and the 9th international joint conference on natural language processing (EMNLP-IJCNLP)},
  pages={3982--3992},
  year={2019}
}

@article{pan2026vla,
  title={VLA-Corrector: Lightweight Detect-and-Correct Inference for Adaptive Action Horizon},
  author={Pan, Yi and Pan, Miao and Lu, Qi and Huang, Jiaming and Zhang, Man and Huang, Siteng and Li, Xin and Zhang, Jie and Shen, Yongliang and Zhang, Xuhong and others},
  journal={arXiv preprint arXiv:2607.01804},
  year={2026}
}

@article{sawada2025cernet,
  title={CERNet: Class-Embedding Predictive-Coding RNN for Unified Robot Motion, Recognition, and Confidence Estimation},
  author={Sawada, Hiroki and Pitti, Alexandre and Quoy, Mathias},
  journal={arXiv preprint arXiv:2512.07041},
  year={2025}
}

@article{annabi2022continual,
  title={Continual sequence modeling with predictive coding},
  author={Annabi, Louis and Pitti, Alexandre and Quoy, Mathias},
  journal={Frontiers in Neurorobotics},
  volume={16},
  pages={845955},
  year={2022},
  publisher={Frontiers Media SA}
}

@article{annabi2021bidirectional,
  title={Bidirectional interaction between visual and motor generative models using predictive coding and active inference},
  author={Annabi, Louis and Pitti, Alexandre and Quoy, Mathias},
  journal={Neural Networks},
  volume={143},
  pages={638--656},
  year={2021},
  publisher={Elsevier}
}

@inproceedings{lin2026evo,
  title={Evo-1: Lightweight vision-language-action model with preserved semantic alignment},
  author={Lin, Tao and Zhong, Yilei and Du, Yuxin and Zhang, Jingjing and Liu, Jiting and Chen, Yinxinyu and Gu, Encheng and Liu, Ziyan and Cai, Hongyi and Zou, Yanwen and others},
  booktitle={Proceedings of the IEEE/CVF conference on computer vision and pattern recognition},
  pages={13397--13406},
  year={2026}
}

@article{wen2025tinyvla,
  title={TinyVLA: toward fast, data-efficient vision-language-action models for robotic manipulation},
  author={Wen, Junjie and Zhu, Yichen and Li, Jinming and Zhu, Minjie and Tang, Zhibin and Wu, Kun and Xu, Zhiyuan and Liu, Ning and Cheng, Ran and Shen, Chaomin and others},
  journal={IEEE Robotics and Automation Letters},
  volume={10},
  number={4},
  pages={3988--3995},
  year={2025},
  publisher={IEEE}
}

@article{brohan2023rt,
  title={Rt-2: Vision-language-action models transfer web knowledge to robotic control},
  author={Brohan, Anthony and Brown, Noah and Carbajal, Justice and Chebotar, Yevgen and Chen, Xi and Choromanski, Krzysztof and Ding, Tianli and Driess, Danny and Dubey, Avinava and Finn, Chelsea and others},
  journal={arXiv preprint arXiv:2307.15818},
  year={2023}
}

@article{team2024octo,
  title={Octo: An open-source generalist robot policy},
  author={Team, Octo Model and Ghosh, Dibya and Walke, Homer and Pertsch, Karl and Black, Kevin and Mees, Oier and Dasari, Sudeep and Hejna, Joey and Kreiman, Tobias and Xu, Charles and others},
  journal={arXiv preprint arXiv:2405.12213},
  year={2024}
}

@article{kim2024openvla,
  title={Openvla: An open-source vision-language-action model},
  author={Kim, Moo Jin and Pertsch, Karl and Karamcheti, Siddharth and Xiao, Ted and Balakrishna, Ashwin and Nair, Suraj and Rafailov, Rafael and Foster, Ethan and Lam, Grace and Sanketi, Pannag and others},
  journal={arXiv preprint arXiv:2406.09246},
  year={2024}
}

@article{black2024pi_0,
  title={$pi\_0 $: A Vision-Language-Action Flow Model for General Robot Control},
  author={Black, Kevin and Brown, Noah and Driess, Danny and Esmail, Adnan and Equi, Michael and Finn, Chelsea and Fusai, Niccolo and Groom, Lachy and Hausman, Karol and Ichter, Brian and others},
  journal={arXiv preprint arXiv:2410.24164},
  year={2024}
}

@article{kim2025fine,
  title={Fine-tuning vision-language-action models: Optimizing speed and success},
  author={Kim, Moo Jin and Finn, Chelsea and Liang, Percy},
  journal={arXiv preprint arXiv:2502.19645},
  year={2025}
}

@article{shukor2025smolvla,
  title={Smolvla: A vision-language-action model for affordable and efficient robotics},
  author={Shukor, Mustafa and Aubakirova, Dana and Capuano, Francesco and Kooijmans, Pepijn and Palma, Steven and Zouitine, Adil and Aractingi, Michel and Pascal, Caroline and Russi, Martino and Marafioti, Andres and others},
  journal={arXiv preprint arXiv:2506.01844},
  year={2025}
}

@article{chi2025diffusion,
  title={Diffusion policy: Visuomotor policy learning via action diffusion},
  author={Chi, Cheng and Xu, Zhenjia and Feng, Siyuan and Cousineau, Eric and Du, Yilun and Burchfiel, Benjamin and Tedrake, Russ and Song, Shuran},
  journal={The International Journal of Robotics Research},
  volume={44},
  number={10-11},
  pages={1684--1704},
  year={2025},
  publisher={Sage Publications Sage UK: London, England}
}

@article{rao1999predictive,
  title={Predictive coding in the visual cortex: a functional interpretation of some extra-classical receptive-field effects},
  author={Rao, Rajesh PN and Ballard, Dana H},
  journal={Nature neuroscience},
  volume={2},
  number={1},
  pages={79--87},
  year={1999},
  publisher={Nature Publishing Group}
}

@article{friston2010free,
  title={The free-energy principle: a unified brain theory?},
  author={Friston, Karl},
  journal={Nature reviews neuroscience},
  volume={11},
  number={2},
  pages={127--138},
  year={2010},
  publisher={Nature Publishing Group UK London}
}

@inproceedings{hwang2017predictive,
  title={Predictive coding-based deep dynamic neural network for visuomotor learning},
  author={Hwang, Jungsik and Kim, Jinhyung and Ahmadi, Ahmadreza and Choi, Minkyu and Tani, Jun},
  booktitle={2017 Joint IEEE International Conference on Development and Learning and Epigenetic Robotics (ICDL-EpiRob)},
  pages={132--139},
  year={2017},
  organization={IEEE}
}

@article{choi2018generating,
  title={Generating goal-directed visuomotor plans based on learning using a predictive coding-type deep visuomotor recurrent neural network model},
  author={Choi, Minkyu and Matsumoto, Takazumi and Jung, Minju and Tani, Jun},
  journal={arXiv preprint arXiv:1803.02578},
  year={2018}
}

@article{ahmadi2019novel,
  title={A novel predictive-coding-inspired variational RNN model for online prediction and recognition},
  author={Ahmadi, Ahmadreza and Tani, Jun},
  journal={Neural computation},
  volume={31},
  number={11},
  pages={2025--2074},
  year={2019},
  publisher={MIT Press One Rogers Street, Cambridge, MA 02142-1209, USA journals-info~…}
}

@article{wang2026world,
  title={World action models: The next frontier in embodied ai},
  author={Wang, Siyin and Shi, Junhao and Fu, Zhaoyang and He, Xinzhe and Liu, Feihong and Yang, Chenchen and Zhou, Yikang and Fei, Zhaoye and Gong, Jingjing and Fu, Jinlan and others},
  journal={arXiv preprint arXiv:2605.12090},
  year={2026}
}

@article{zhang2026world,
  title={From World Models to World Action Models: A Concise Tutorial for Robotics},
  author={Zhang, Xiaoxiong and Zeng, Xiong and Zhang, Wei},
  journal={arXiv preprint arXiv:2607.00836},
  year={2026}
}

@inproceedings{he2016deep,
  title={Deep residual learning for image recognition},
  author={He, Kaiming and Zhang, Xiangyu and Ren, Shaoqing and Sun, Jian},
  booktitle={Proceedings of the IEEE conference on computer vision and pattern recognition},
  pages={770--778},
  year={2016}
}

@article{zhao2023learning,
  title={Learning fine-grained bimanual manipulation with low-cost hardware},
  author={Zhao, Tony Z and Kumar, Vikash and Levine, Sergey and Finn, Chelsea},
  journal={arXiv preprint arXiv:2304.13705},
  year={2023}
}

@article{bu2025univla,
  title={Univla: Learning to act anywhere with task-centric latent actions},
  author={Bu, Qingwen and Yang, Yanting and Cai, Jisong and Gao, Shenyuan and Ren, Guanghui and Yao, Maoqing and Luo, Ping and Li, Hongyang},
  journal={arXiv preprint arXiv:2505.06111},
  year={2025}
}

@article{idei2026predictive,
  title={Predictive processing as a scalable computational principle for embodied multitask intelligence},
  author={Idei, Hayato and Miyake, Tamon and Ogata, Tetsuya and Yamashita, Yuichi},
  journal={Science Advances},
  volume={12},
  number={33},
  pages={eaed7511},
  year={2026},
  publisher={American Association for the Advancement of Science}
}

@article{yamashita2008emergence,
  title={Emergence of functional hierarchy in a multiple timescale neural network model: a humanoid robot experiment},
  author={Yamashita, Yuichi and Tani, Jun},
  journal={PLoS computational biology},
  volume={4},
  number={11},
  pages={e1000220},
  year={2008},
  publisher={Public Library of Science San Francisco, USA}
}

@article{liu2026oa,
  title={Oa-wam: Object-addressable world action model for robust robot manipulation},
  author={Liu, Yushan and Sun, Peibo and Li, Shoujie and Xie, Yifan and Zhang, Lingfeng and Chao, Xintao and Dong, Shiyuan and Chen, Fang and Zhang, Xiao-Ping and Ding, Wenbo},
  journal={arXiv preprint arXiv:2605.06481},
  year={2026}
}

@article{liu2023libero,
  title={Libero: Benchmarking knowledge transfer for lifelong robot learning},
  author={Liu, Bo and Zhu, Yifeng and Gao, Chongkai and Feng, Yihao and Liu, Qiang and Zhu, Yuke and Stone, Peter},
  journal={Advances in Neural Information Processing Systems},
  volume={36},
  pages={44776--44791},
  year={2023}
}

@inproceedings{choi2017predictive,
  title={Predictive coding for dynamic vision: Development of functional hierarchy in a multiple spatio-temporal scales RNN model},
  author={Choi, Minkyu and Tani, Jun},
  booktitle={2017 International Joint Conference on Neural Networks (IJCNN)},
  pages={657--664},
  year={2017},
  organization={IEEE}
}

@article{cheang2024gr,
  title={Gr-2: A generative video-language-action model with web-scale knowledge for robot manipulation},
  author={Cheang, Chi-Lam and Chen, Guangzeng and Jing, Ya and Kong, Tao and Li, Hang and Li, Yifeng and Liu, Yuxiao and Wu, Hongtao and Xu, Jiafeng and Yang, Yichu and others},
  journal={arXiv preprint arXiv:2410.06158},
  year={2024}
}

@article{hu2024video,
  title={Video prediction policy: A generalist robot policy with predictive visual representations},
  author={Hu, Yucheng and Guo, Yanjiang and Wang, Pengchao and Chen, Xiaoyu and Wang, Yen-Jen and Zhang, Jianke and Sreenath, Koushil and Lu, Chaochao and Chen, Jianyu},
  journal={arXiv preprint arXiv:2412.14803},
  year={2024}
}

@inproceedings{wang2026unified,
  title={Unified vision-language-action model},
  author={Wang, Yuqi and Li, Xinghang and Wang, Wenxuan and Zhang, Junbo and Li, Yingyan and Chen, Yuntao and Wang, Xinlong and Zhang, Zhaoxiang},
  booktitle={International Conference on Learning Representations},
  volume={2026},
  pages={80929--80944},
  year={2026}
}

@article{zhu2025unified,
  title={Unified world models: Coupling video and action diffusion for pretraining on large robotic datasets},
  author={Zhu, Chuning and Yu, Raymond and Feng, Siyuan and Burchfiel, Benjamin and Shah, Paarth and Gupta, Abhishek},
  journal={arXiv preprint arXiv:2504.02792},
  year={2025}
}

\end{document}